\pdfoutput=1

\documentclass[11pt]{article}

\usepackage{acl}

\usepackage{times}
\usepackage{latexsym}

\usepackage[T1]{fontenc}

\usepackage[utf8]{inputenc}

\usepackage{microtype}

\usepackage{multirow}
\usepackage{booktabs}
\usepackage{adjustbox}
\usepackage{hyperref}

\usepackage{algorithm} 
\usepackage{subfigure}
\usepackage{graphicx}
\usepackage{algpseudocode} 
\usepackage{amsmath}  

\usepackage{amssymb}

\newcommand{\entpair}[2]{$\langle\,e_{#1},e_{#2}\,\rangle$}

\newcommand{\relto}[1]{\overset{{#1}}{\longrightarrow}}
\newcommand{\modelname}{{MERIt}}  

\newcommand{\ntgtents}{{target entities}}
\newcommand{\ntgtentp}{{target entity pair}}

\title{\modelname: \underline{Me}ta-Path Guided Contrastive Learning for Logical \underline{R}eason\underline{i}ng}

\author{Fangkai Jiao$^\dag$, Yangyang Guo${^\S}$\footnotemark[1],~ Xuemeng Song$^\dag$, Liqiang Nie${^\dag}$\footnotemark[1] \\
  $^\dag$School of Computer Science and Technology, Shandong University, Qingdao, China \\
  $^\S$School of Computing, National University of Singapore \\
  {\small\texttt{jiaofangkai@hotmail.com}}
    \quad {\small\texttt{\{guoyang.eric, sxmustc, nieliqiang\}@gmail.com}}
}

\begin{document}
\maketitle

\renewcommand{\thefootnote}{\fnsymbol{footnote}}
\footnotetext[1]{Corresponding author: Yangyang Guo and Liqiang Nie.}
\renewcommand{\thefootnote}{\arabic{footnote}}

\begin{abstract}
Logical reasoning is of vital importance to natural language understanding.
Previous studies either employ graph-based models to incorporate prior knowledge about logical relations, or introduce symbolic logic into neural models through data augmentation. 
These methods, however, heavily depend on annotated training data, and thus suffer from over-fitting and poor generalization problems due to the dataset sparsity.
To address these two problems, in this paper, we propose \modelname, a \underline{ME}ta-path guided contrastive learning method for logical \underline{R}eason\underline{I}ng of text, to perform self-supervised pre-training on abundant unlabeled text data.
Two novel strategies serve as indispensable components of our method. 
In particular, a strategy based on meta-path is devised to discover the logical structure in natural texts, followed by a counterfactual data augmentation strategy to eliminate the information shortcut induced by pre-training. 
The experimental results on two challenging logical reasoning benchmarks, i.e., ReClor and LogiQA, demonstrate that our method outperforms the SOTA baselines with significant improvements.\footnote{Our code and pre-trained models are available at \url{https://github.com/SparkJiao/MERIt}.}
\end{abstract}

\section{Introduction}

\begin{figure}[t]
    \centering
    \includegraphics[width=1.0\linewidth]{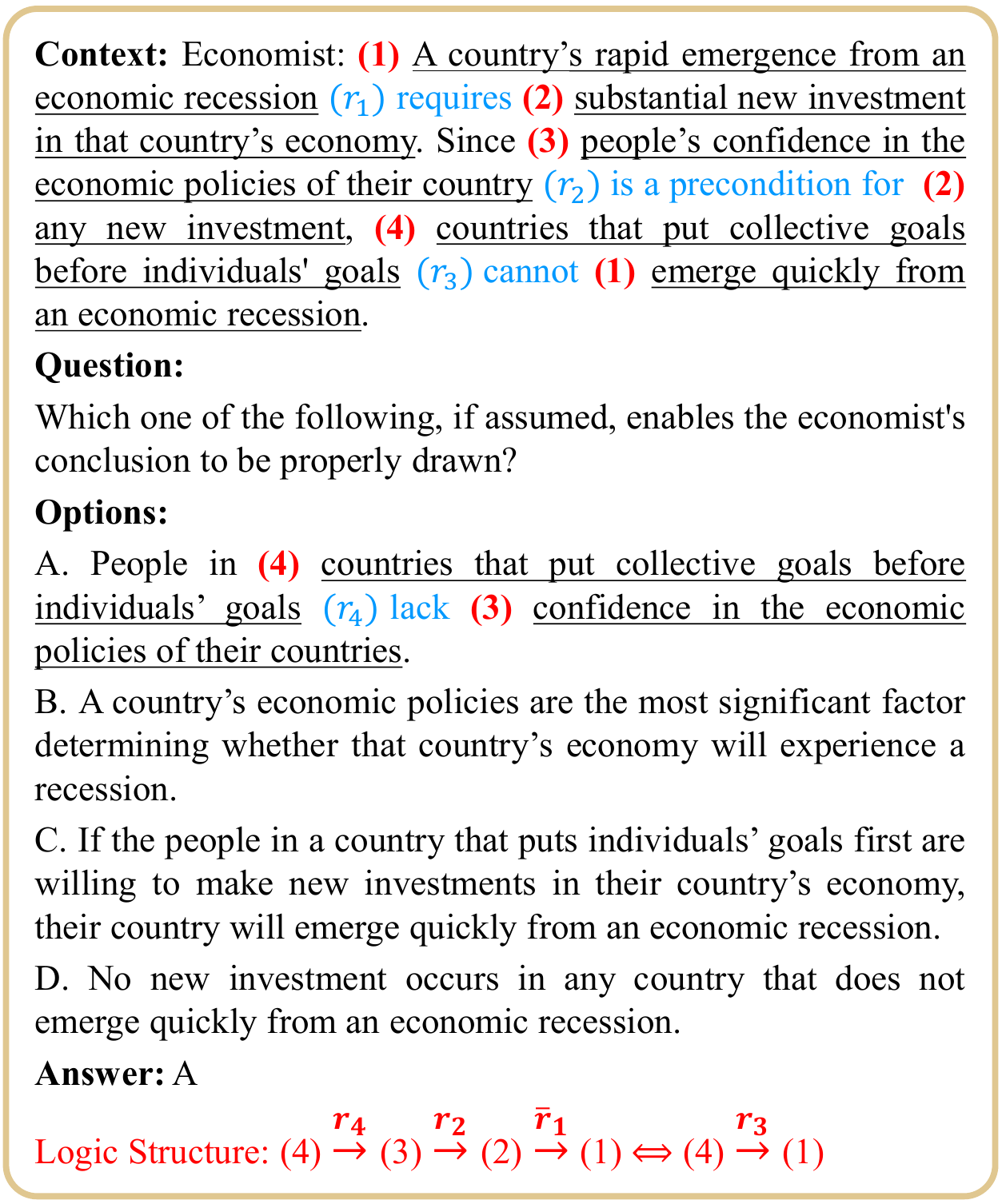}
    \caption{An instance of logical reasoning from the ReClor dataset. 
    To infer the right answer, we should uncover the underlying logical structure, as shown in the bottom. 
    (x) represents the logical variable (e.g., entity or phrase) and $r_j$ denotes the relation (e.g., predicate) between two logical variables. $\bar{r}_j$ is the passive relation of $r_j$.}
    \label{fig:example}
    \vspace{-0.7cm}
\end{figure}

Logical reasoning has long been recognized as one key critical thinking ability of human being.
Until very recently, some pioneer researchers have crystallized this for the NLP community, and built several public challenging benchmarks, such as ReColor~\citep{reclor} and LogiQA~\citep{logiqa}. 
Logical reasoning\footnote{We refer the term \emph{logical reasoning} to the task itself in the remaining of this paper.} requires to correctly infer the semantic relations with respect to the constituents among different sentences. 
A typical formulation of logical reasoning is illustrated in Figure~\ref{fig:example}, namely, a real-world examination instance from ReClor.
As can be seen, to find the correct answer for the given question, one needs to extract the logical structures residing in a pair of each option and the whole context,
and justify its reasonableness.

As a matter of fact, logical reasoning is still at its initial stage, thence, existing studies are somewhat rare in literature.
Some efforts have been devoted to designing specific model architectures or integrating symbolic logic as the hints attached to the potential logical structure.
For instance, 
\citet{DAGN} and \citet{focal-reasoner} first constructed a graph of different constituents and then performed implicit reasoning with graph neural networks (GNNs).
\citet{lreasoner} proposed LReasoner, a unified context extension and data augmentation framework based on the parsed logical expressions.
These approaches have achieved some progress on benchmark datasets. 
However, though equipped with pre-trained language models, they still suffer from problems like overfitting and poor generalization.
We attribute these drawbacks to the difficulty of building a model aware of the logical relations beneath natural language, which is revealed from two sides: 1) 
the high sparsity of the existing datasets, 
and 2) the goal of general pre-training, i.e., masked language modeling~\citep{bert}, which however, deviates largely from that of the logical reasoning. 
To tackle this issue, we aim to 
build a bridge between logical reasoning and self-supervised pre-training, and accordingly inherit the strong generalization power from pre-trained language models.
Our proposed method is inspired by the recent progress of contrastive learning
based pre-training. 
It mainly consists of two novel components: meta-path guided data construction and counterfactual data augmentation.
Both components are leveraged to perform automatic instance composition from unlabeled corpus (e.g., Wikipedia) for contrastive learning.
Regarding the first component, we propose to employ the meta-path to define a symbolic form of logical structure.
The intuition behind this is that the logical structure can be expressed as a reasoning path composed of a series of relation triplets, and a meta-path inherently offers such a means of  consistency~\citep{knowledge-logic-eswc2021}.
Specifically, given an arbitrary document and a pair of entities in it,
we try to find a positive instance pair in the document according to the logical structure.
And the negative ones can thus be generated by modifying the relations involved in the structure, which explicitly break the logical consistency.
Nevertheless, the contrastive learning often fails
when models easily locate trivial solutions~\citep{shortcut-mrc-2021-lai}.
In this context, the pre-trained language model may exclude the negative options through their conflicts with the world knowledge.
To eliminate this information shortcut, in our second novel component, we devise a strong counterfactual data augmentation~\citep{counterfactual-ner} strategy.
By mixing counterfactual data during pre-training, of which the positive instance pair is also against the world knowledge, this component shows more advantage in reasoning over logical relations.

We integrate this method with both ALBERT~\citep{albert} and RoBERTa~\citep{roberta}\footnote{In this paper, we refer ALBERT-xxlarge and RoBERTa-large to \textit{ALBERT} and \textit{RoBERTa} for simplicity, respectively.} for further pre-training, and then fine-tune them on two downstream logical reasoning benchmarks, i.e., ReClor and LogiQA.
The experimental results demonstrate that our method can outperform all the existing strong baselines, yet without any augmentation from the original training data. 
Besides, the ablation studies also show the effectiveness of the two essential strategies in our method.
The contribution of this paper is summarized as follows:
\begin{enumerate}
    \item We propose \modelname, a MEta-path guided contrastive learning method for logical ReasonIng of text, to reduce the heavy reliance on annotated data.
    To the best of our knowledge, we are the first to explore  self-supervised pre-training for logical reasoning. 
    \item We successfully employ the meta-path strategy to mine the potential logical structure in raw text. It is able to automatically generate negative candidates for contrastive learning via logical relation editing.
    \item We propose a simple yet effective counterfactual data augmentation method to eliminate the information shortcut during pre-training.
    \item We evaluate our method on two logical reasoning tasks, LogiQA and ReClor. The experimental results show that our method achieves the new state-of-the-art performance on two benchmark datasets.
\end{enumerate}

\label{sec:intro}

\section{Related Work}

\subsection{Self-Supervised Pre-training}

With the success of language modeling based pre-training~\citep{bert,gpt}, designing self-supervised pretext tasks to facilitate specific downstream ones has been extensively studied thus far.
For example,
\citet{realm} proposed to train the retriever jointly with the encoder via retrieval enhanced masked language modeling for open-domain question answering.
\citet{jiao2021rept} devised a retrieval-based pre-training approach to bridge the gap between language modeling and machine reading comprehension by enhancing the evidence extraction ability.
\citet{reason-bert} proposed ReasonBERT to facilitate complex reasoning over multiple and hybrid contexts.
The model is pre-trained on automatically constructed query-evidence pairs, which involve different types of corpora and long-range relations.

In addition, contrastive learning~\citep{cl2006lecun} contributes to a strong toolkit to implement self-supervised pre-training. 
The key to contrastive learning is to build efficacious positive and negative counterparts.
For example, 
\citet{gao2021simcse} leveraged Dropout~\citep{dropout} 
to build positive pairs from the same sentence while keeping the semantics untouched.
Other sentences in the same mini-batch serve as negative candidates to obtain better sentence embeddings.
ERICA~\citep{erica} is a knowledge enhanced language model pre-trained through entity and relation discrimination, where the negative candidates are sampled from the pre-defined dictionaries.
Nevertheless, directly employing these contrastive learning approaches to logical reasoning is arduous. 
One possible reason to this is 
the absence of distant labels or strong assumptions to group the naturally occurring text by its logical structure.

\vspace{-0.1cm}
\subsection{Logical Reasoning}

Logical reasoning has attracted increasing research attention recently.
Devising specific model architectures and integrating symbolic logic have been proved to be two effective solutions.
For example,
\citet{DAGN} and \citet{focal-reasoner} proposed to extract the basic units for logical reasoning, e.g., the elementary discourse or fact units, and then employed GNNs to model  possible relationships.
The graph structure of constituents can be viewed as a form of prior knowledge pertaining to logical relations.
Differently, \citet{critical-thinking-lm} and \citet{transformer-as-soft-reasoner} used synthetically generated datasets to prove that the Transformer~\citep{transformer} or pre-trained GPT-2 is able to perform complex reasoning, motivating  following researchers to introduce symbolic rules into neural models.
For example,
\citet{lreasoner} developed a context extension and data augmentation framework, which is based on the extracted logical expressions.
Superior performance over its contenders can be observed on the ReClor dataset.


In this paper, we propose a self-supervised contrastive learning approach to enhance the logical reasoning ability of neural models. 
Orthogonal to existing methods, our approach is endowed with two intriguing merits:
1) it shows strong advantage in utilizing the unlabeled text data,
and 2) the symbolic logic is seamlessly introduced into neural models via the guidance of meta-path for automatic data construction.


\section{Preliminary}
\subsection{Contrastive Learning}
\label{sec:pre:cl}

Contrastive Learning (CL) aims to learn recognizable representations by pulling the semantically similar examples close and pushing apart the dissimilar ones~\citep{cl2006lecun}.
Given an instance $x$, a semantically similar example $x^+$, and a set of dissimilar examples $\mathcal{X}^-$ to $x$, the objective of CL can be formulated as:
\begin{equation}
    \begin{aligned}
    \mathcal{L}_{\mathrm{CL}}&=L(x, x^+, \mathcal{X}^-) \\
                            &=-\log\frac{\exp f(x, x^+)}{\sum_{x'\in\mathcal{X}^-\cup\{x^+\}}\exp f(x,x')}
    \end{aligned}
    \label{eqn:cl-obj}
\end{equation}
where $f$ is the model to be optimized.


\subsection{Symbolic Logical Reasoning}

As shown in Figure~\ref{fig:example}, given a context containing a series of logical variables $\{v_1,v_2,\cdots,v_n\}$, and the relations between them, the logical reasoning objective is to judge whether a triplet $\langle\,v_i,r_{i,j},v_j\,\rangle$ in language, where $r_{i,j}$ is the relation between $v_i$ and $v_j$, can be inferred from the context through a reasoning path:
\begin{equation}
    \langle\,v_i,r_{i,j},v_j\,\rangle\gets(v_i\relto{r_{i,i+1}} v_{i+1}\cdots\relto{r_{j-1,j}}v_{j}).
    \label{eqn:logic-rule}
\end{equation}
The equation is also referred to \textit{symbolic logic rules}~\citep{transformer-as-soft-reasoner,knowledge-logic-eswc2021}.

\subsection{Meta-Path}
\label{sec:pre:mp}

Given an entity-level knowledge graph, where the nodes refer to entities and edges are the relations among them, the meta-path connecting two \ntgtents~\entpair{i}{j} can be given as,
\begin{equation}
    e_i\relto{r_{i,i+1}}e_{i+1}\relto{r_{i+1,i+2}}\cdots e_{j-1}\relto{r_{j-1,j}}e_j,
    \label{eqn:path}
\end{equation}
where $r_{i,j}$ denotes the relation between entities $e_i$ and $e_j$. 
The meta-path in the entity-level knowledge graph are often employed as a particular data structure expressing the relation between two indirectly connected entities~\citep{dou-graph-doc-re,discriminative-reasoning-for-doc-re}.

\begin{figure*}[t]
    \centering
    \includegraphics[width=1.0\textwidth]{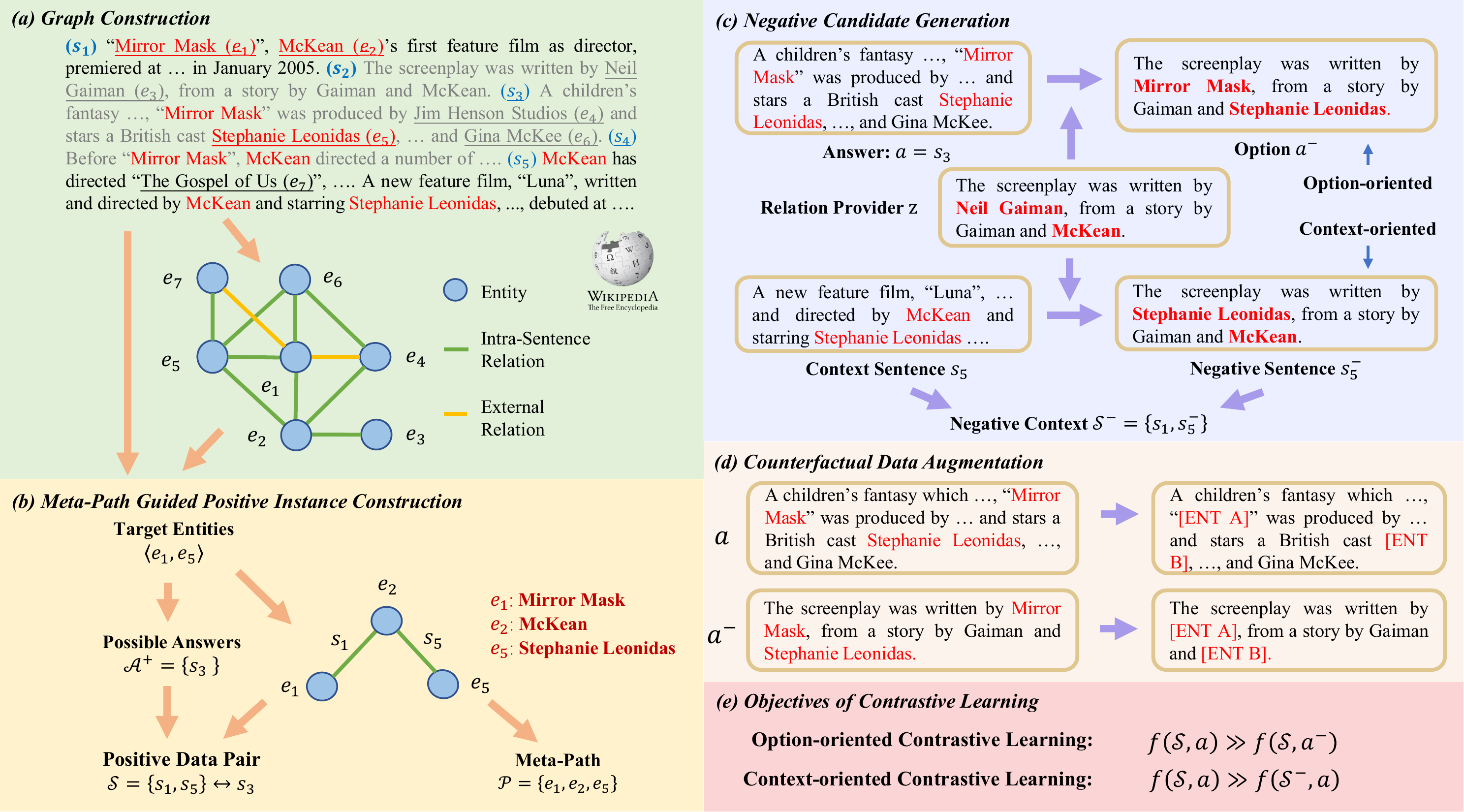}
    \caption{The overall framework of our proposed method. 
    (a) A document $\mathcal{D}$ from Wikipedia and the corresponding entity-level graph construction. The sentences in black will be extracted as the context input for (b).
    (b) Given two \ntgtents~\entpair{1}{5}, the possible answers $\mathcal{A}^+$ and the meta-path are firstly extracted. 
    The context sentences $\mathcal{S}$ connecting the entities in the meta-path, and the answers in $\mathcal{A}$, are leveraged to yield positive instance pairs.
    (c) Given a sentence $z$ with alternative relations, the relation modification for negative context sentence and option construction is implemented through entity replacement. The top operation is performed for negative options while the bottom one is to facilitate negative contexts.
    (d) The counterfactual sentences are generated by entity replacement to eliminate the information shortcut during pre-training.
    (e) The generated positive and negative samples are used for contrastive learning.
    }
    \label{fig:model-framework}
    \vspace{-0.6cm}
\end{figure*}



\section{Method}

In this paper, we study the problem of logical reasoning on the task of multiple choice question answering (MCQA). 
Specifically, given a passage $P$, a question $Q$ and a set of $K$ options $\mathcal{O}=\{O_1,\cdots,O_K\}$, the goal is to select the correct option $O_y$, where $y\in[1,K]$.
Notably, to tackle this task, we devise a novel pre-training method equipped with contrastive learning, where the abundant knowledge contained in the large-scale Wikipedia documents is explored.
We then transfer the learned knowledge to the downstream logical reasoning task.

\subsection{From Logical Reasoning to Meta-Path}

In a sense, in MCQA for logical reasoning, both the given context (i.e., passage and question) and options express certain relations between different logical variables (Figure~\ref{fig:example}). 
Go a step further, following Equation~\ref{eqn:logic-rule}, the relation triplet contained in the correct option should be deduced from the given context through a reasoning path,
while that in the wrong options should not. 
In other words, the context is logically consistent with the correct option only.



In light of this, the training instances for our contrastive learning based pre-training should be in the form of a context-option pair, where the context consists of multiple sentences and expresses the relations between the included constituents, while the option should illustrate the potential relations between parts of the constituents.  
Nevertheless, it is non-trivial to derive such instance pairs from large-scale unlabeled corpus like Wikipedia due to the redundant constituents, e.g., nouns and predicates.
In order to address it, we propose to take the entities contained in unlabeled text as logical variables, and Equation~\ref{eqn:logic-rule} can be transformed as:
\begin{equation}
    \langle\,e_{i},r_{{i},{j}},e_{j}\,\rangle \gets (e_i\relto{r_{i,i+1}} e_{i+1} \cdots \relto{r_{j-1,j}}e_j).
    \label{eqn:ent-logic-rule}
\end{equation}
As can be seen, the right part above is indeed a meta-path connecting \entpair{i}{j} as formulated in Equation~\ref{eqn:path}, indicating an indirect relation between \entpair{i}{j} through intermediary entities and relations.
In order to aid the logical consistency conditioned on entities to be established, we posit an  assumption that
\textit{under the same context (in the same passage), the definite relation between a  pair of entities can be inferred from the contextual indirect one, or at least not logically contradict to it}.
Taking the passage in Figure~\ref{fig:model-framework} as an example, 
it can be concluded from the sentences $s_1$ and $s_5$ that, the director \textit{McKean} has cooperated with \textit{Stephanie Leonidas}. 
Therefore, the logic is consistent between $\{s_1,s_5\}$ and $s_3$.
This can be viewed as a weaker constraint than the original one in Equation~\ref{eqn:logic-rule} for logical consistency, yet it can be further enhanced by constructing negative candidates violating logics.

Motivated by this, given an arbitrary document $\mathcal{D}=\{s_1,\cdots,s_m\}$, where $s_i$ is the $i$-th sentence,
we can first build an entity-level graph, denoted as $\mathcal{G}=(\mathcal{V},\mathcal{E})$, where $\mathcal{V}$ is the set of entities contained in $\mathcal{D}$ and $\mathcal{E}$ denotes the set of relations between entities. 
Notably, to comprehensively capture the relations among entities, we take into account both the external relation from the knowledge graph and the intra-sentence relation. As illustrated in Figure~\ref{fig:model-framework} (a), there will be an intra-sentence relation between two entities if they are mentioned in a common sentence.
Thereafter, we can derive the pre-training instance pairs according to the meta-paths extracted from the graph, which will be detailed in the following subsections.






\subsection{Meta-Path Guided Positive Instance Construction}
\label{sec:method:meta-path}

As defined in Equation~\ref{eqn:ent-logic-rule}, in the positive instances, the answer should contain a relation triplet that is logically consistent with the given context. 
Since we take the intra-sentence relationship into consideration, given a pair of entities contained in the document,
we first collect the sentences mentioning both of them as the set of answer candidates.
Accordingly, we then try to find a meta-path connecting the entity pair and hence derive the corresponding logically consistent context. 

In particular, as shown in Figure~\ref{fig:model-framework} (b), given an entity pair \entpair{i}{j}, we denote the collected answer candidates as $\mathcal{A}^+$,
and then we use Depth-First Search~\citep{dfs-Tarjan72} to find a meta-path linking them on $\mathcal{G}$, following Equation~\ref{eqn:path}.
Thereafter, the context sentences $\mathcal{S}$ corresponding to the answer candidates in $\mathcal{A}^+$ are derived by retrieving those sentences undertaking the intra-sentence relations during the search algorithm.
Finally, for each answer candidate $a\in\mathcal{A}^+$, the pair $(\mathcal{S}, a)$ is treated as a positive context-answer pair to facilitate our contrastive learning.
The details of positive instance generation algorithm are described in Appendix~\ref{sec:appendix:dfs-alg}.

\subsection{Negative Instance Generation}
\label{sec:method:data-construction}


In order to obtain the negative instances (i.e., negative context-option pairs) where the option is not logically consistent with the context,
the most straightforward way is to randomly sample the sentences from different documents.
However, this approach could lead to trivial solutions by simply checking whether the entities involved in each option are the same as those in the given context.
In the light of this, we resort to directly breaking the logical consistency of the positive instance pair by modifying the relation rather than the entities in the context or the option, to derive the negative instance pair. 

In particular, given a positive instance pair $(\mathcal{S}, a)$, we devise two negative instance generation methods: the context-oriented and the option-oriented method, focusing on generating negative pairs by modifying the relations involved in the context $\mathcal{S}$ and answer $a$ of the positive pair, respectively.
Considering that the relation is difficult to be extracted, especially the intra-sentence relation,
we propose to implement this reversely via the entity replacement.
In particular, for the option-oriented method, suppose that \entpair{i}{j} is the \ntgtentp~ for retrieving the answer $a$, 
we first randomly sample a sentence $z$ that contains at least one different entity pair \entpair{a}{b} from \entpair{i}{j} as the relation provider. 
We then obtain the negative option by replacing the entities $e_a$ and $e_b$ in $z$ with $e_i$ and $e_j$, respectively.
The operation is equivalent to replacing the relation contained in $a$ with that in $z$. Formally, we denote the operation as 
$$
a^-=\mathrm{Relation\_Replace}(z\to a).
$$

Pertaining to the context-oriented negative instance generation method, we first 
randomly sample a sentence $s_i\in\mathcal{S}$, and then conduct the modification process as follows,
$$
s_i^-=\mathrm{Relation\_Replace}(z\to s_i),
$$
where the entity pair to be replaced in $s_i$ should be contained in the meta-path corresponding to the \ntgtentp~\entpair{i}{j}.
Accordingly, the negative context can be written as $\mathcal{S}^-=\mathcal{S}\setminus\{s_i\}\cup\{s_i^-\}$.
Figure~\ref{fig:model-framework}~(c) illustrates the above operations on both the answer and context sentence.

\subsection{Counterfactual Data Augmentation}
\label{sec:method:counterfactual}

According to~\citet{position-bias-2020-ko,sigir-bias,shortcut-mrc-2021-lai,tip-bias}, the neural models are adept at finding a trivial solution through the illusory statistical information in datasets to make correct predictions,
which often leads to inferior generalization.
In fact, this issue can also occur in our scenario.
In particular, since the correct answer is from a natural sentence and describes a real world fact, while the negative option is synthesized by entity replacement, which may conflict with the commonsense knowledge.
As a result, the pre-trained language model tends to identify the correct option directly by judging its factuality rather than the logical consistency with the given context.
For example, as shown in Figure~\ref{fig:model-framework} (d) (left), 
the language model deems $a$ as correct, simply due to that the other synthetic option $a^-$ conflicts with the world knowledge.

To overcome this problem, we develop a simple yet effective counterfactual data augmentation method to further improve the capability of logical reasoning~\citep{counterfactual-ner}. 
Specifically, given the entities $\mathcal{P}$ that are involved in the meta-path, we randomly select some entities from $\mathcal{P}$ and replace their occurrences in the context and the answer of the positive instance pair $(\mathcal{S}, a)$ with the entities extracted from other documents. In this manner, the positive instance also contradicts to the world knowledge.
Notably, considering that the positive and negative instance pairs should keep the same set of entities, we also conduct the same replacement for $a^-$ or $\mathcal{S}^-$, if they mention the selected entities.
As illustrated in Figure~\ref{fig:model-framework} (d) (right), a counterfactual instance can be generated by replacing \textit{Mirror Mask} and \textit{Stephanie Leonidas} in $a$ and $a^-$ with \textbf{[ENT A]} and \textbf{[ENT B]}, where \textbf{[ENT A]} and \textbf{[ENT B]} are arbitrary entities.
Ultimately, the key to infer the correct answer lies in the accurate inference of the logical relation between entities \textbf{[ENT A]} and \textbf{[ENT B]} implied in each context-option pair.
We provide more cases of the constructed data and their corresponding counterfactual samples in Appendix~\ref{sec:appendix:data-case}.


\subsection{Contrastive Learning based Pre-training}
\label{sec:method:cl}

As discussed in previous subsection, there are 
two contrastive learning schemes: option-oriented CL and context-oriented CL. 
Let $\mathcal{A}^-$ be the set of all constructed negative options with respect to the correct option $a$. The option-oriented CL can be formulated as:
\begin{equation}
    \mathcal{L}_{\mathrm{OCL}}=L(\mathcal{S},a,\mathcal{A}^-).
    \label{eqn:ocl}
\end{equation}
In addition, given $\mathcal{C}^-$ as the set of all generated negative contexts corresponding to $\mathcal{S}$, the objective of context-oriented CL can be written as:
\begin{equation}
    \mathcal{L}_{\mathrm{CCL}}=L(a,\mathcal{S},\mathcal{C}^-).
    \label{eqn:ccl}
\end{equation}
To avoid the catastrophic forgetting problem, we also add the MLM objective during pre-training and the final loss is:
\begin{equation}
\mathcal{L}=\mathcal{L}_{\mathrm{OCL}} + \mathcal{L}_{\mathrm{CCL}}+\mathcal{L}_{\mathrm{MLM}}.
\end{equation}

\begin{figure}
    \centering
    \includegraphics[width=0.9\linewidth]{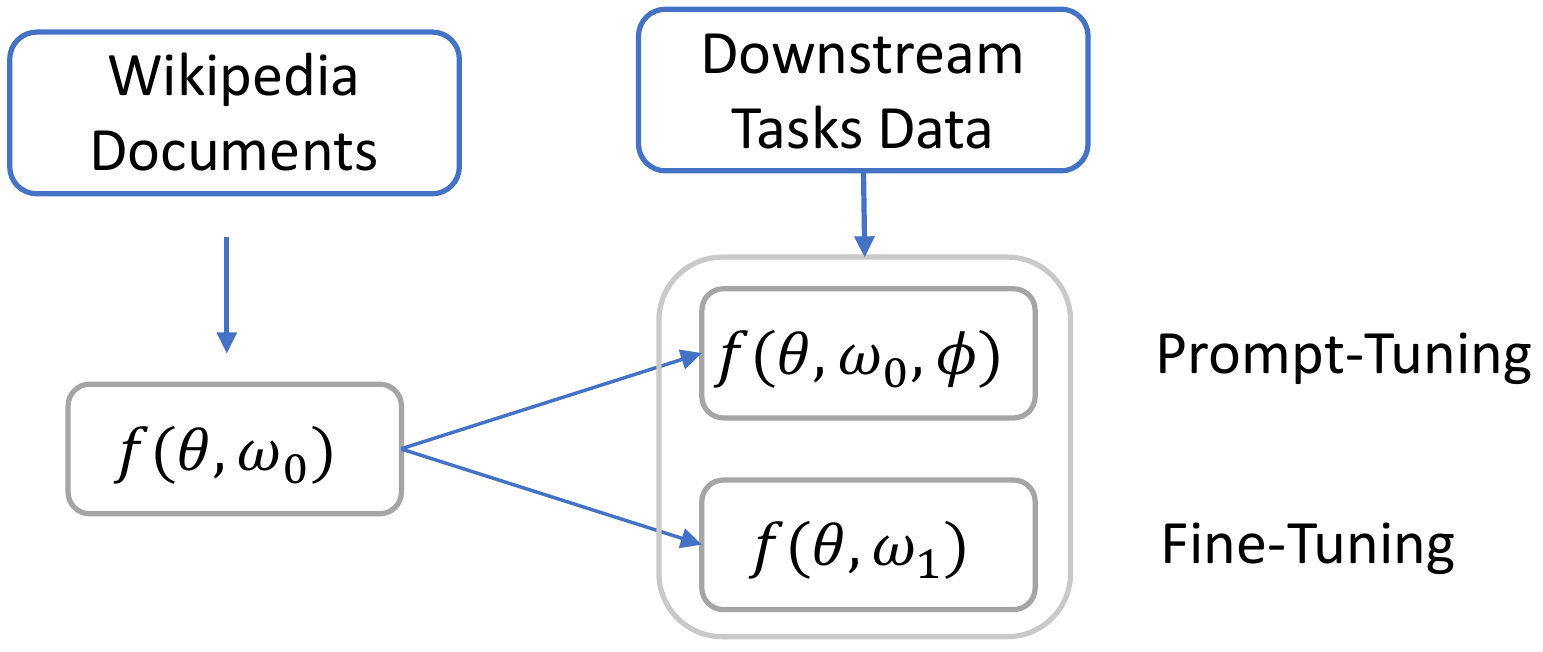}
    \caption{The overall training scheme of our method.}
    \label{fig:training_frame}
    \vspace{-0.5cm}
\end{figure}

\begin{table*}[t]
\centering
\setlength{\tabcolsep}{7.0mm}{
\scalebox{0.75}{
\begin{tabular}{lcccccc}
\toprule
\multicolumn{1}{c}{\multirow{2}{*}{Model / Dataset}} & \multicolumn{4}{c}{\textbf{ReClor}}                                                                                    & \multicolumn{2}{c}{\textbf{LogiQA}}                        \\
\multicolumn{1}{c}{}                                 & \multicolumn{1}{c}{Dev} & \multicolumn{1}{c}{Test} & \multicolumn{1}{c}{Test-E} & \multicolumn{1}{c}{Test-H} & \multicolumn{1}{c}{Dev} & \multicolumn{1}{c}{Test} \\ \hline
RoBERTa                                         & 62.6                     & 55.6                     & 75.5                       & 40.0                        & 35.0                    & 35.3                        \\
DAGN                                                  & 65.2                     & 58.2                     & 76.1                       & 44.1                        & 35.5                    & 38.7                    \\ 
DAGN (Aug)                                            & 65.8                     & 58.3                     & 75.9                       & 44.5                        & 36.9                    & 39.3                     \\ 
LReasoner (RoBERTa)$^\ddag$                     & 64.7                     & 58.3                     & 77.6                       & 43.1                        & ---                     & ---                      \\
Focal Reasoner                                        & 66.8                     & 58.9                     & 77.1                       & 44.6                        & \textbf{41.0}           & 40.3        \\ 
\hline
\modelname                                        & 66.8                     & 59.6                     & 78.1                       & 45.2                        & 40.0                  & 38.9                 \\
\modelname~+ LReasoner                            & 67.4                     & 60.4                     & 78.5                       & 46.2                        & ---                   & ---                  \\
\modelname~+ Prompt                             & \textbf{69.4}            & \textbf{61.6}            & 79.3              & \textbf{47.8}               & 39.9                  & \textbf{40.7}         \\
\modelname~+ Prompt + LReasoner                       & 67.3                     & 61.4                      & \textbf{79.8}      & 46.9                        & ---                      & --- \\
\hline
ALBERT                                        & 69.1 & 66.5 & 76.7 & 58.4 & 38.9 & 37.6 \\
\modelname~(ALBERT)                           & 74.2 & 70.1 & 81.6 & 61.0 & 43.7 & \textbf{42.5} \\
\modelname~(ALBERT) + Prompt                  & \textbf{74.7} & \textbf{70.5} & \textbf{82.5} & \textbf{61.1} & \textbf{46.1} & 41.7 \\
\hline
\textit{max} \\ \hline
LReasoner (RoBERTa)                         & 66.2                     & 62.4                     & 81.4                       & 47.5                        & 38.1               & 40.6                      \\ 
\modelname                                       & 67.8 & 60.7 & 79.6 & 45.9 & \textbf{42.4} & 41.5  \\
\modelname~+ Prompt                             & \textbf{70.2}            & \textbf{62.6}            & 80.5                       & \textbf{48.5}               & 39.5      & \textbf{42.4}            \\
\hline
LReasoner (ALBERT)                            & 73.2 & 70.7 & 81.1 & 62.5 & 41.6 & 41.2 \\
\modelname~(ALBERT)                           & 73.2 & 71.1 & 83.6 & 61.3 & 43.9 & \textbf{45.3} \\
\modelname~(ALBERT) + Prompt                  & \textbf{75.0} & \textbf{72.2} & \textbf{82.5} & \textbf{64.1} & \textbf{45.8} & 43.8 \\
\bottomrule
\end{tabular}
}}
\caption{The overall results on ReClor and LogiQA. We adopt the \textbf{accuracy} as the evaluation metric and all the baselines are based on RoBERTa except specific statement. For each model we repeated training for 5 times using different random seeds and reported the average results. 
$^\ddag$: The results are reproduced by ourselves.
\textit{max}: The results of the model achieving the best accuracy on the test set.}
\label{tab:overall-results}
\vspace{-0.5cm}
\end{table*}

\vspace{-0.2cm}
\subsection{Fine-tuning}
\label{sec:method:fine-tuning}

During the fine-tuning stage, to approach the task of MCQA, we adopt the following loss function:
\begin{equation}
\mathcal{L}_{\mathrm{QA}}=-\log\frac{\exp{f(P,Q,O_y)}}{\sum_i\exp{f(P,Q,O_i)}},
\label{eqn:qa}
\end{equation}
where $O_y$ is the ground-truth option for the question $Q$, given the passage $P$.

Figure~\ref{fig:training_frame} shows the overall training scheme of our method. $f$ is the model to be optimized, $\theta$, $\omega_0$, $\omega_1$ and $\phi$ are parameters of different modules. 
During pre-training, we use a 2-layer MLP as the output layer. The parameters of the output layer are denoted as $\omega_0$, and $\theta$ represents the pre-trained Transformer parameters.
As for the fine-tuning stage, we employ two schemes. 
For simple fine-tuning, we follow~\citet{bert} to add another 2-layer MLP with randomly initialized parameters $\omega_1$ on the top of the pre-trained Transformer.
In addition, to fully take advantage the knowledge acquired during pre-training stage, we choose to directly fine-tune the pre-trained output layer with optimizing both $\theta$ and $\omega_0$.
In order to address the discrepancy that the question is absent during pre-training, the prompt-tuning technique~\citep{prompt-tuning} is employed.
Specifically, some learnable embeddings with randomly initialized parameters $\phi$ are appended to the input to transform the question in downstream tasks into declarative constraint.

\section{Experiment}

\subsection{Dataset and Baseline}
We evaluated our method on two challenging logical reasoning benchmarks, i.e., LogiQA and ReClor, with several strong baselines, including the pre-trained language models, DAGN~\citep{DAGN}, Focal Reasoner~\citep{focal-reasoner} and LReasoner~\citep{lreasoner}.
For more details, please refer to Appendix~\ref{sec:appendix:experiment}.

\subsection{Implementation Detail}

We further pre-trained RoBERTa and ALBERT on Wikipedia for another 500 and 100 steps, respectively, and the batch size for pre-training is set to 4,096. All experiments conducted on downstream tasks are repeated for 5 times with different random seeds.
The knowledge graph we used for constructing training data is provided by ~\citet{erica}. 
More implementation details can be found in Appendix~\ref{sec:appendix:implement}.



\section{Result and Analysis}


\subsection{Overall Results}

The overall results on ReClor and LogiQA are shown in Table~\ref{tab:overall-results}.
It can be observed that 
1) \modelname~outperforms all the strong baselines using the same backbone with significant improvements.
Besides, our method achieves the new state-of-the-art performance on both datasets.
2) Our method leads to drastic contribution to the original models without further pre-training, i.e., RoBERTa and ALBERT, and the prompt-tuning further enhances our model with a significant performance margin, which both demonstrate the potential of our pre-training method.
3) \modelname~achieves better performance on the more difficult split of ReClor (Test-H), indicating that our pre-training method is less affected by the statistical shortcut~\citep{reclor}.
4) \modelname~+ Prompt does not benefit from the framework of LReasoner significantly.
This is probably because the basic knowledge about logic rules has been covered in our method.
5) We also report the best result on the test set on LogiQA and ReClor for fair comparison with the published results of LReasoner. 
It can be observed that in terms of the best accuracy on the test set, our model still outperforms LReasoner consistently based on both RoBERTa and ALBERT.





\subsection{Ablation Study}

Table~\ref{tab:ablation} shows the results of our ablation studies.
To observe the impacts brought by the meta-path strategy, we built a baseline model without the meta-path strategy by
randomly selecting the sentences in a passage to form the context-answer pairs.

From this table we can conclude that: 
1) the model without counterfactual data augmentation (- DA) has a severe performance degradation.
It suggests that the counterfactual data is essential for \modelname~to conduct logical reasoning. 
As for the ratio of original data to the counterfactual one, on test set,
we found that 1:3 (+ DA$^3$) leads to better performance using prompt tuning while 1:2 (+ DA$^2$) obtains the best performance using simple fine-tuning.
2) The model without the guidance of meta-path (- Meta-Path) demonstrates a much worse performance than \modelname, 
indicating that the meta-path strategy plays an important role by discovering the potential logic structure.
3) Considering the results of models without the objectives of option-oriented CL and context-oriented CL, it can be seen that both contrastive learning schemes are beneficial for logical reasoning.
In addition, the context-oriented CL is more effective than option-oriented CL.
One possible reason to this is that the context-oriented CL is more diverse in format since each sentence can be disturbed while the option-oriented CL will make the model pay more attention to the option, 
leading to a worse generalization during fine-tuning.




\begin{table}[t]
\centering
\setlength{\tabcolsep}{0.9mm}{
\scalebox{0.9}{
\begin{tabular}{lcccc}
\toprule
Model                   & Dev  & Dev (P.) & Test  & Test (P.) \\ \hline
\modelname              & 66.8 & 69.4     & 59.6  & 61.6      \\
-~DA                   & 63.0 & 64.5     & 57.9  & 59.8      \\
+~DA$^2$               & 65.3 & 67.8       & \textbf{60.2}  & 61.3      \\
+~DA$^3$                & 66.2 & 68.0     & 59.3  & \textbf{61.9}       \\ 
-~Option-oriented CL    & 63.8 & 65.4     & 58.9 & 61.5      \\
-~Context-oriented CL   & 64.0 & 66.5     & 58.8 & 60.2      \\
-~Meta-Path           & 64.8 & 65.1     & 58.0  & 60.8       \\
\bottomrule
\end{tabular}
}}
\caption{Performance comparisons on ReClor between different variants of \modelname. \textit{DA} means data augmentation
and \textit{DA}$^N$ refers to 1:N ratio of the original data to the augmented data. 
\textit{P.} is short for \textit{Prompt Tuning}.}
\label{tab:ablation}
\vspace{-0.2cm}
\end{table}

\begin{figure}[t]
    \centering
    \includegraphics[width=1.0\linewidth]{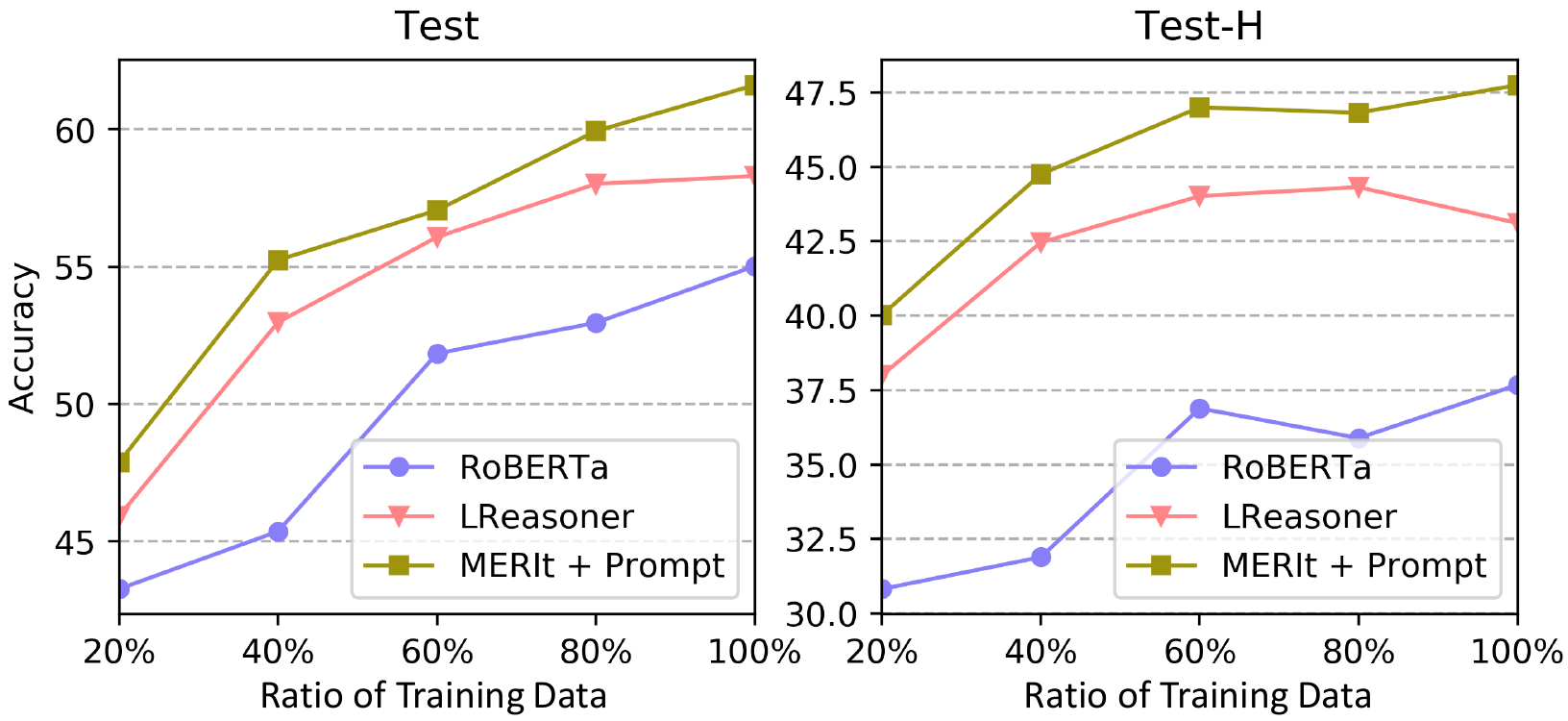}
    \caption{Results on the test set (left) and the test-H set (right) of ReClor.}
    \label{fig:data-size}
    \vspace{-0.5cm}
\end{figure}

\subsection{Performance with Limited Training Data}

Figure~\ref{fig:data-size} shows the accuracy on the test set and test-H set of ReClor with respect to different amount of training data.
We reported the average results of \modelname~+ Prompt, LReasoenr and RoBERTa.
It can be observed that:
1) With the scale of training data becoming larger, the performance of all models achieves improvements.
2) \modelname~+ Prompt shows better performance under low resource, especially on test-H.
Our method trained on 40\% data has achieved comparable performance with RoBERTa.
In addition, on test-H, our method outperforms RoBERTa and LReasoner trained on full dataset using only 20\% and 40\% training data, respectively,
evidently demonstrating the generalization capability of our method.
3) Further improvements to LReasoner become insignificant when consuming more training data.
This suggests that the basic logic rules can be easily fitted.

\begin{figure}[t]
    \centering
    \includegraphics[width=1.0\linewidth]{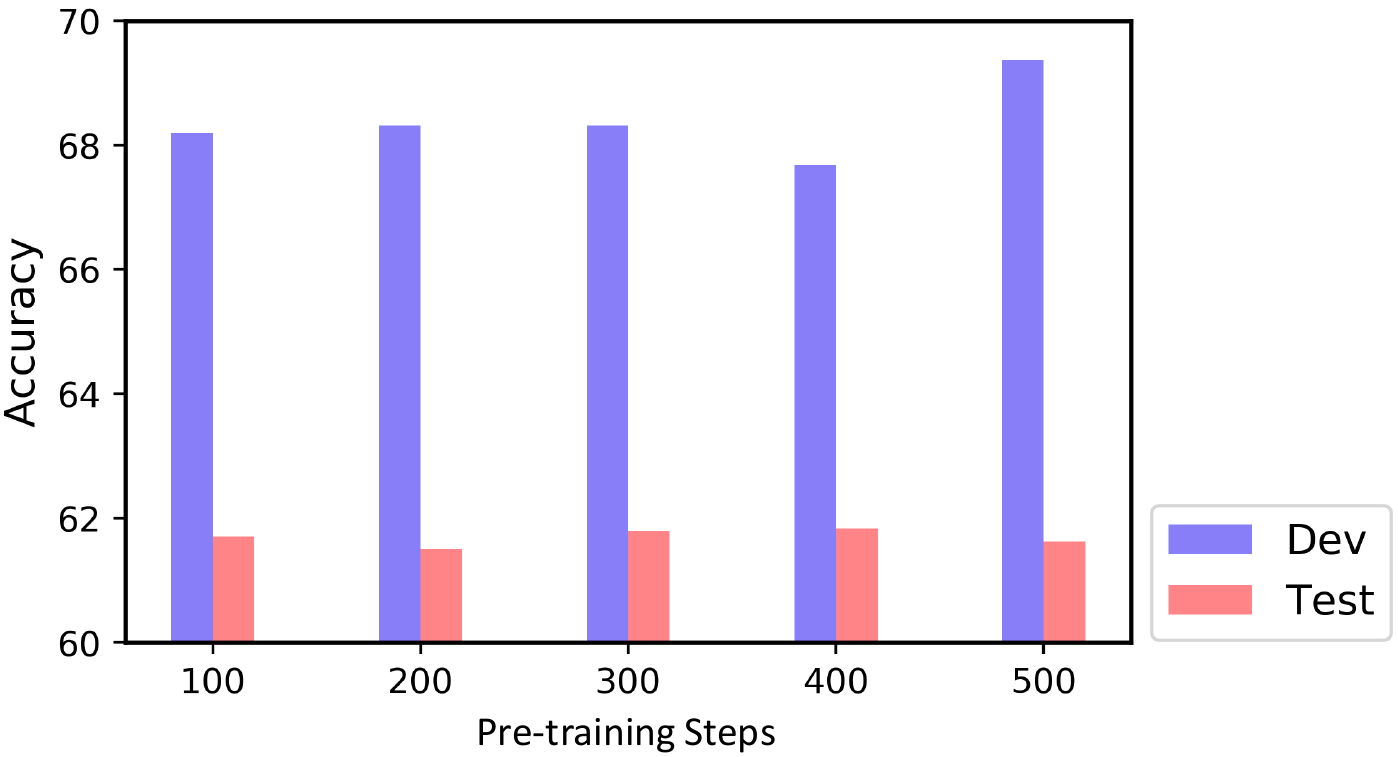}
    \caption{The prompt-tuning results on ReClor using the models pre-trained with different steps.}
    \label{fig:step-ablation}
\end{figure}

\begin{table}[t]
\centering
\setlength{\tabcolsep}{5.0mm}{
\scalebox{1.0}{
\begin{tabular}{lcc}
\toprule
Model                   & Dev  & Test  \\ \hline
RoBERTa                 & 84.9 & 84.2  \\ 
\modelname              & \textbf{85.9} & \textbf{85.5}  \\

\bottomrule
\end{tabular}
}}
\caption{The accuracy of different models on DREAM dataset.}
\label{tab:dream}
\vspace{-0.2cm}
\end{table}

\begin{table}[t]
\centering
\setlength{\tabcolsep}{0.9mm}{
\scalebox{0.85}{
\begin{tabular}{lcccc}
\toprule
Model                           & Dev  & Test & Test-E  & Test-H \\ \hline
DeBERTa-v2-xlarge               & 76.7 & 71.0 & 83.8    & 60.9  \\ 
\modelname~(DeBERTa-v2-xlarge)  & \textbf{78.0} & \textbf{73.1} & \textbf{86.2} & \textbf{64.4}  \\ \hline
DeBERTa-v2-xxlarge              & 78.3 & 75.3 & 84.0    & 68.4 \\
\modelname~(DeBERTa-v2-xxlarge) & \textbf{80.6} & \textbf{78.1} & \textbf{84.6}    & \textbf{72.9} \\

\bottomrule
\end{tabular}
}}
\caption{Results on ReClor with DeBERTa as the backbone.}
\label{tab:deberta}
\vspace{-0.2cm}
\end{table}



\subsection{Effect of Pre-training Steps}

In order to explore the effects of pre-training steps, we fine-tuned the models pre-trained for different steps on ReClor and the results are shown in Figure~\ref{fig:step-ablation}. 
From the histogram we can find that our method achieves the best performance on dev set at 500 steps.
Besides, the model pre-trained with 100 steps (using only around 410k samples) has achieved comparable performance with the best one, 
indicating that our method is very competitive with few training iterations.

\subsection{Performance on DREAM}

We also evaluated our method on another benchmark requiring complex reasoning abilities, DREAM~\citep{dream}, to verify its generalization ability to different tasks. As shown in Table~\ref{tab:dream}, our method can also make significant improvements compared with RoBERTa, demonstrating the generalization ability of our method.

\subsection{Results of DeBERTa}

Table~\ref{tab:deberta} shows the results of DeBERTa-v2-xlarge and DeBERTa-v2-xxlarge on ReClor, which validate that our method can be scaled to stronger pre-trained language models with significant improvements.








\section{Conclusion and Future Work}

In this paper, we present \modelname, a meta-path guided contrastive learning method to facilitate logical reasoning via self-supervised pre-training.
\modelname~is built upon the meta-path strategy for automatic data construction and the counterfactual data augmentation to eliminate the information shortcut during pre-training.
With the evaluation on two logical reasoning benchmarks, our method has obtained significant improvements over strong baselines relying on task-specific model architecture or augmentation of original dataset.
Pertaining to the further work, we plan to strengthen our method from both data construction and model architecture design angles.
More challenging instances are expected to be constructed if multiple meta-paths can be considered at the same time.
Besides, leveraging GNNs may bring better interpretability and generalization since the graph structure can be integrated into both pre-training and fine-tuning stages.

\section*{Acknowledgements}
We sincerely appreciate the valuable comments from all the reviewers to help us make the paper polished.
We also greatly thank to Liqiang Jing and Harry Cheng for their kind suggestions. 
This work is supported by the National Natural Science Foundation of China, No.:U1936203; the Shandong Provincial Natural Science Foundation, No.:ZR2019JQ23; and Young creative team in universities of Shandong Province, No.:2020KJN012.

\bibliography{anthology,custom}
\bibliographystyle{acl_natbib}

\clearpage
\appendix
\section{DFS-based Algorithm for Meta-Path Extraction}
\label{sec:appendix:dfs-alg}
\vspace{-0.2cm}

\begin{algorithm}[thb]
  \caption{The DFS algorithm to obtain the meta-paths.}  
  \label{alg:dfs}
  \begin{algorithmic}[1]  
    \Require
      The graph $\mathcal{G}=(\mathcal{E}, \mathcal{V})$;
      The sentences of the document $\mathcal{D}=\{s_1,\cdots,s_m\}$;
      The entity set of the $i$-th sentence $\mathcal{V}_i$;
    \Ensure  
      $\mathcal{P}$, $\mathcal{S}$, and $\mathcal{A}^+$;
    
    \For{each $(e_i,e_j)\in\mathcal{V}\times\mathcal{V}$ and $i\neq j$}
        \State $\mathcal{A}^+=\{s_k|e_i\in\mathcal{V}_k,e_j\in\mathcal{V}_k\}$;
        \State $\mathcal{D}'=\mathcal{D}\setminus\mathcal{A}^+$;
        \State $\mathrm{cond},\mathcal{P},\mathcal{S}\gets\mathrm{DFS}(e_i,\{e_i\},\varnothing,e_j,\mathcal{G},\mathcal{D}')$;
        \If{cond is TRUE and $\mathcal{A}^+$ is not $\varnothing$}
            \State \Return $\mathcal{A}^+,\mathcal{P},\;\mathcal{S}$;
        \EndIf
    \EndFor
    \State \Return $\varnothing,\varnothing$, $\varnothing$;
    \State 
    \Function {DFS}{$e_i,\mathcal{P}',\mathcal{S}',e_d, \mathcal{G}=(\mathcal{E},\mathcal{V}),\mathcal{D}'$}
        \If{$e_i=e_d$}
            \State \Return TRUE, $\mathcal{P}',\mathcal{S}'$;
        \EndIf
        \For{each $(e_j,s_k)\in\mathcal{V}\times\mathcal{D}'$ and $(e_i,e_j)\in\mathcal{E},e_j\in\mathcal{V}_k$}
                \State $\mathcal{G}'=(\mathcal{E},\mathcal{V}\setminus\{e_j\})$;
                \State $\mathcal{P}''=\mathcal{P}'\cup\{e_j\}$;
                \If{$e_i\in\mathcal{V}_k$}
                    \State $\mathcal{D}''=\mathcal{D}'\setminus\{s_k\}$;
                    \State $\mathcal{S}''=\mathcal{S}'\cup\{s_k\}$;
                \Else
                    \State $\mathcal{D}''=\mathcal{D}',\mathcal{S}''=\mathcal{S}'$;
                \EndIf
                \State \Return DFS($e_j,\mathcal{P}'',\mathcal{S}'',e_d,\mathcal{G}',\mathcal{D}''$);
        \EndFor
        \State \Return FALSE, $\varnothing$, $\varnothing$;
    \EndFunction
    
  \end{algorithmic}
\end{algorithm}
\vspace{-0.2cm}

\section{Details of Experimental Setup}
\label{sec:appendix:experiment}

\subsection{Dataset}
\label{sec:appendix:dataset}

\noindent\textbf{ReClor}~\citep{reclor} is extracted from logical reasoning questions of standardized graduate admission examinations.
The held-out test set is further divided into EASY and HARD subsets, denoted as test-E and test-H, respectively.
The instances in test-E are biased and can be solved even without knowing contexts and questions by neural models.
A leaderboard\footnote{\url{https://eval.ai/web/challenges/challenge-page/503/leaderboard/1347}.} is also host for public evaluation.

\noindent\textbf{LogiQA}~\citep{logiqa} consists of 8,678 multiple-choice questions collected from National Civil Servants Examinations of China and are manually translated into English by experts. 
The dataset is randomly split into train/dev/test sets with 7,376/651/651 samples, respectively.
LogiQA contains various logical reasoning types, e.g., categorical reasoning and sufficient conditional reasoning.

\subsection{Baseline}
\label{sec:appendix:baseline}

\noindent\textbf{DAGN}~\cite{DAGN} is a discourse-aware graph network that reasons on the discourse structure of texts.
It is based on elementary discourse units and discourse relations. DAGN (Aug) is a variant that augments the graph features.

\noindent\textbf{Focal Reasoner}~\cite{focal-reasoner} is a fact-driven logical reasoning model, which builds super-graphs on the top of fact units as the basis for logical reasoning.
It captures both global connections between facts and the local concepts or actions inside the fact.

\noindent\textbf{LReasoner}~\cite{lreasoner} includes a context extension framework and a data augmentation algorithm, which are all conducted based on the extracted logical expressions. 
This method has achieved new state-of-the-art performance on ReClor recently.

Besides, we also compare the performance with the directly fine-tuned large pre-trained language models, including RoBERTa and ALBERT.

\section{Implementation Detail}
\label{sec:appendix:implement}

\subsection{Data Construction}
\label{sec:appendix:data-construct}

During the data construction process, we have employed two tricks to improve the complexity of the pretext task:
\begin{enumerate}
    \item For the sentence $z$ as the relation provider for negative instance construction, the sentences from the document are primarily to be considered because they share the same entities with the context or describe the same topic.
    This can also be viewed as a trick to avoid trivial solution by checking whether the samples come from the same domain.
    Another problem is that if $z$ comes from the same document, taking the option-oriented method as example, the replacement may not work if $e_i=e_a$ and $e_j=e_b$.
    To address it, we will change the order of the entities to be replaced, i.e., swapping the mentions of $e_i$ and $e_j$.
    \item Similarly, for counterfactual data augmentation, 
    supposing the extracted meth-path of a training instance connects an entity pair \entpair{i}{j}, $e_i$ and $e_j$ are always considered to be replaced for generating counterfactual data.
    And thus the sets of answer candidates $\mathcal{A}^+$ constructed from other documents, where the corresponding meta-paths also link \entpair{i}{j}, can be employed as negative candidates directly.
    The motivation of the trick is to avoid modifications on the original texts as many as possible.
\end{enumerate}

\subsection{Pre-training Setting}

We employed the model implementation of Transformer from Huggingface~\citep{huggingface-tf} and pytorch\footnote{\url{https://pytorch.org}.} framework.
The corpus for pre-training is generated from the dataset provided by \citet{erica}\footnote{\url{https://github.com/thunlp/ERICA}.}, which includes the pre-processed passages from Wikipedia and the recognized entities with their distantly annotated relations. 
The generated corpus contains one million samples and each sample has 3 negative options. 

During pre-training, we adopted the LAMB~\citep{lamb-optim} optimizer, warming up the learning rate to the peak and then linearly decaying it. It takes 32 hours on 4 RTX 2080Ti GPUs for RoBERTa pre-training and 3 days on 2 TeslaT4 GPUs for ALBERT pre-training. Other hyper-parameters for pre-training are reported in Table~\ref{tab:params-pt}. 

\begin{table}[t]
\centering
\scalebox{0.85}{
\begin{tabular}{lcc}
\toprule
                             & \textbf{ALBERT}            & \textbf{RoBERTa}\\ \hline
Batch Size                   & 4096                       & 4096   \\
Peak Learning Rate           & 5e-5                       & 1e-4   \\
Training Steps               & 100                        & 500                       \\
Warmup Proportion            & 0.2                        & 0.1                       \\
Weight Decay                 & 0.01                       & 0.01                       \\
Adam $\epsilon$              & 1e-6                       & 1e-6                      \\
Adam $\beta_1$               & 0.9                        & 0.9                       \\
Adam $\beta_2$               & 0.98                       & 0.98                       \\
Max Sequence Length          & 256                        & 320                      \\
Gradient Clipping            & 5.0                        & 5.0                       \\ 
Hidden Size of MLP           & 8192                       & 2048 \\
\bottomrule
\end{tabular}
}
\caption{Hyper-parameters for ALBERT and RoBERTa during pre-training, respectively.}
\label{tab:params-pt}
\vspace{-0.2cm}
\end{table}


\subsection{Hyper-parameters for Fine-tuning}
\label{sec:appendix:param}
The random seeds we utilized for repeated experiments are 42, 43, 44, 45 and 4321.
The hyper-parameters for fine-tuning are shown in Table~\ref{tab:params-ft}.

\section{Case Study for Generated Examples}
\label{sec:appendix:data-case}
\begin{figure*}[t]
    \centering
    \includegraphics[width=1.0\textwidth]{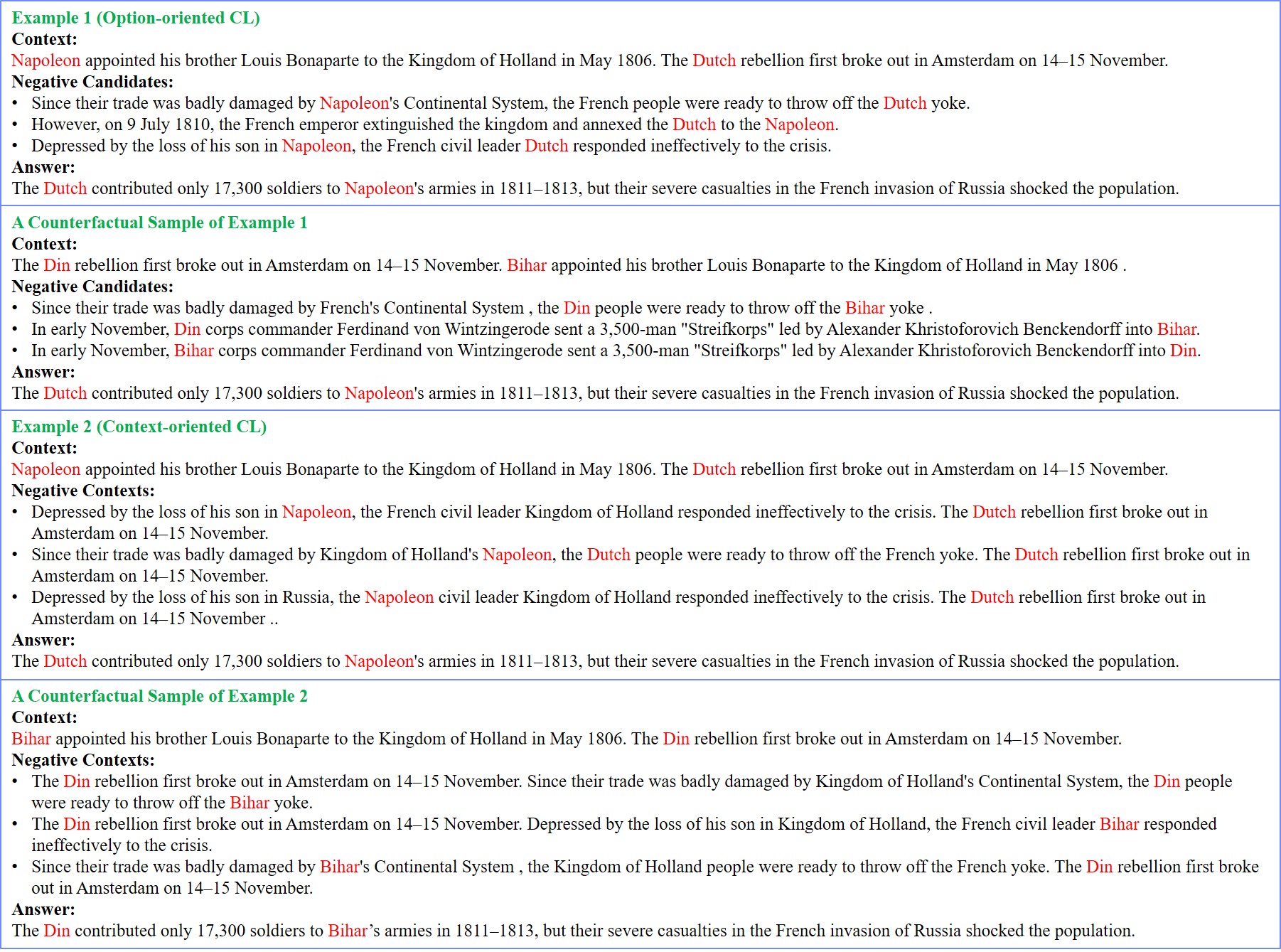}
    \caption{Two cases of the generated and the counterfactual examples. The \ntgtents~used for extracting meta-path are colored in red.}
    \label{fig:data-case}
\end{figure*}

Figure~\ref{fig:data-case} shows the constructed examples for contrastive learning as well as the corresponding counterfactual examples.

\begin{table}[t]
\centering
\setlength{\tabcolsep}{0.9mm}{
\scalebox{1.0}{
\begin{tabular}{lcccc}
\toprule
Model                   & Dev  & Test & Test-E  & Test-H \\ \hline
RoBERTa                 & 35.8 & 35.7 & 44.5    & 28.8  \\ 
\modelname~(500 steps)  & \textbf{39.0} & 35.2 & 41.8    & 30.0  \\
~~100 steps             & 37.5 & \textbf{38.1} & \textbf{47.5}    & 30.6 \\
~~200 steps             & 38.1 & 38.0 & 47.3    & 30.7 \\
~~300 steps             & 37.4 & 36.4 & 43.6    & 30.7 \\
~~400 steps             & 38.5 & 35.9 & 42.5    & 30.7 \\ \hline
ALBERT                  & 43.6 & 40.2 & 46.6    & 35.2 \\
\modelname~(ALBERT)     & \textbf{46.3} & \textbf{44.6} & \textbf{51.8}    & \textbf{38.9} \\

\bottomrule
\end{tabular}
}}
\caption{Results of Linear Probing on ReClor.}
\label{tab:linear-probing}
\vspace{-0.2cm}
\end{table}

\section{Results for Linear Probing}

Table~\ref{tab:linear-probing} shows the results of linear probing on ReClor, where we used a single linear layer as the output layer and only fine-tuned its parameters.
As shown in the table, \modelname~(100 steps) and \modelname~(ALBERT) outperform RoBERTa and ALBERT on both dev and test set, respectively.

\section{A Different View from Contrastive Graph Representation Learning}
\label{sec:appendix:gcl}

To understand why the pre-training approach can promote logical reasoning, we provide a different view from the contrastive learning for graphs.
Following \citet{gcl-gcc-2020}, $x$ and $x^+$ in Equation~\ref{eqn:cl-obj} are different sub-graphs extracted from the same graph through random walk with restart~\citep{random-walk-restart} while $x^-$ is sub-graph sampled from a different graph.
To avoid the trivial solution by simply checking whether the node indices of two sub-graphs match, they also developed an anonymization operation by relabeling the nodes of each sub-graph.
In fact, our proposed method can be taken as a special case of graph contrastive learning.
Firstly, the context and answer based on the meta-path can be viewed as sub-graphs of $\mathcal{G}$.
In particular, the answer is the sub-graph with only two nodes (the two entities connected by the meta-path).
Secondly, the entity replacement for negative candidates construction and counterfactual data generation play similar roles with the anonymization operation.
Both of them aim at guiding the model focus on the logical/graph structure.
The only assumption our approach built upon is that inferring the consistency defined in Equation~\ref{eqn:ent-logic-rule} is in demand of logical reasoning, which has already been explored in many studies for document-level relation extraction~\citep{logic-reason-doc-re-zeng,dou-graph-doc-re}.

\begin{table*}[t]
\centering
\setlength{\tabcolsep}{4.0mm}{
\begin{tabular}{lcccccc}
\toprule
\multicolumn{1}{c}{\multirow{2}{*}{}}   & \multicolumn{2}{c}{\textbf{ALBERT}}                     & \multicolumn{2}{c}{\textbf{RoBERTa}}                        \\
\multicolumn{1}{c}{}                    & \multicolumn{1}{c}{ReClor} & \multicolumn{1}{c}{LogiQA} & \multicolumn{1}{c}{ReClor} & \multicolumn{1}{c}{LogiQA} \\ \hline
Batch Size                              & 24 & 24 & 24 & 16   \\
Peak Learning Rate                      & 2e-5$\clubsuit$/3e-5 & 2e-5 & 1e-5$\clubsuit$/1.5e-5$\spadesuit$ & 8e-6   \\
Epoch                                   & 10 & 10 & 10                         & 10                       \\
Warmup Proportion                       & 0.1 & 0.1 & 0.1                        & 0.2                       \\
Weight Decay                            & 0.01 & 0.01 & 0.01                       & 0.01                       \\
Adam $\epsilon$                         & 1e-6 & 1e-6 & 1e-6                       & 1e-6                      \\
Adam $\beta_1$                          & 0.9 & 0.9 & 0.9                        & 0.9                       \\
Adam $\beta_2$                          & 0.98 & 0.98 & 0.98                       & 0.98                       \\
Max Sequence Length                     & 256$\clubsuit$/231$\spadesuit$ & 256$\clubsuit$/231$\spadesuit$ & 256$\clubsuit$/231$\spadesuit$ & 256$\clubsuit$/231$\spadesuit$ \\
Prefix Length                           & 0$\clubsuit$/5$\spadesuit$ & 0$\clubsuit$/5$\spadesuit$ & 0$\clubsuit$/5$\spadesuit$& 0$\clubsuit$/5$\spadesuit$ \\
Gradient Clipping                       & 0.0 & 0.0 & 0.0                        & 0.0                       \\ 
Dropout                                 & 0.1 & 0.0$\clubsuit$/0.1$\spadesuit$ & 0.1                        & 0.1 \\
\bottomrule
\end{tabular}
}
\caption{Hyper-parameters for fine-tuning on ReClor and LogiQA. $\clubsuit$: Fine-Tuning. $\spadesuit$: Prompt Tuning.}
\label{tab:params-ft}
\end{table*}

\end{document}